# INFERENCE POLICIES

## Paul E. Lehner

Associate Professor
George Mason University
4400 University Drive
Fairfax, VA  22030

Senior Principal Associate
Decision Science Consortium
1895 Preston White Drive
Reston, VA  22071

*Abstract* - In the AI community, there is an ongoing debate as to the most appropriate theory of inferencing under uncertainty. This paper explores the problem of inference from a different perspective. It is suggested that an inferencing system should reflect an *inference policy* that is tailored to the domain of problems to which it is applied -- and furthermore that an inference policy need not conform to any general theory of rational inference or induction. We note, for instance, that Bayesian reasoning about the probabilistic characteristics of an inference domain may result in the specification of a non-Bayesian procedure for reasoning within the inference domain. In this paper, the idea of an inference policy is explored in some detail. To support this exploration, the characteristics of some standard and nonstandard inference policies are examined.

## 1.0 SATISFYING REQUIREMENTS

Consider the following, admitedly artificial, scenario. An inference system must be designed to support a human decision maker. The inference system has only two sources of evidence; degree of belief inputs from expert1 and expert2. For the domain in question, the judgments of both sources are believed to be *reliable*. That is approximately X proportion of inferences believed to degree X are correct (Horwich, 1982). Furthermore, expert1's judgment are generally more extreme than expert2. However, it is uncertain as to the extent to which the two agents judgments are redundant or independent. Since the system must support a human decision maker, it is considered desirable that the inference system also be reliable. Reliability makes it easier for a user to determine circumstances when the aid's advice should be accepted, which often increases the accuracy of the user/machine combination [Lehner, et.al., 1989].

What sort of inference policy will satisfy these requirements? One policy which meets these requirements is simply to ignore expert2 and *routinely* accept the judgments of expert1. From the perspective of most quantitative theories of inference, this is a bad idea -- it routinely ignores obviously useful information. However, since the relationship between the judgments of the two experts is not known, it is not clear how one might merge the two sources of information and still maintain reliability. Consequently, while for *individual* problems within the inference domain this policy seems suboptimal, it may be appropriate for the inference domain *as a whole*.

More generally, *inference policies* should be designed to satisfy a set of requirements determined by examining the anticipated characteristics of an *inference domain*. Often, a standard inference policy (Bayesain, Shaferian, etc.) will satisfy the requirements. Other times, a *nonstandard inference policy* is needed. Below we examine below some standard and nonstandard inference policies to illustrate this approach.

## 2.0 STANDARD INFERENCE POLICIES

We consider below some standard policies, and the types of domain requirements they satisfy. A standard inference policy is defined here as an inference procedure based on any theory of inference seriously considered in the inductive reasoning literature.

### 2.1 Bayesian models

Proponents of the so-called Bayesian approach are generally characertized by their insistence that the *only* rational systems of belief values are point-valued probability models. There are several different lines of argument for this strong assertion. Two of the more popular ones are the Dutchbook and scoring rule arguments. According to the *Dutchbook argument* an agents belief values do not conform to the probability calculus iff there exists a Dutchbook (no-win) gamble that the agent would willingly play. Assuming that ideally rational agents do not accept such gambles, we must conclude that the belief values of such agents conform to the probability calculus. Another line is the *scoring rule argument*. If an agent wishes to minimizes her error rate, and the scoring rule for measuring error is additive. then the *expected* error rate is minimal only if the agents belief values are derived from the probability calculus [Lindley, 1982].

While these arguments may support the Bayesian view of rational induction, they do *not* support the notion that point-valued Bayesian models are necessarily a good inference policy. To show this, we need only point out (as illustrated in Section 1.0) that Bayesian reasoning *about* the characteristics of an inference domain may lead one to conclude that the best inference policy *within* a domain is non-Bayesian. It seems inconsistent for a Bayesian to insist that Bayesian inference is necessarily the best inference policy.

On the other hand, for inference domains that require point-valued estimates, and minimizing error rate seems the appropriate goal, it is hard to imagine how a non-Bayesian system could be appropriate.

### 2.2 Interval Probability Models

The standard litany on expert system (ES) technology claims that ESs encode human expert knowledge. Consequently, a properly engineered ES should make the same inferences that a human expert would. In the knowledge engineering literature, it is considered desirable to base a knowledge base on multiple experts. Consequently, an ES should encode the *common knowledge* of experts and generate belief values that conform with this common knowledge. If experts disagree, then a point-value system cannot possibly reflect common expert knowledge. On the other hand, it is arguable that interval probability systems do maintain common knowledge. If each expert has a goal of minimizing error rate, then each experts belief judgments should conform to the probability calculus. If the knowledge base is composed of interval probability statements that are consistent with each experts judgments, then probability statements derivable from that knowledge base should also be consistent with the judgments of all the experts.

## 2.3 Nonmonotonic Reasoning Logics

Recently, AI researchers have developed a number of formal logics within which it is possible to make defeasible inferences -- categorical inferences that can be later retracted without introducing an inconsistency [Reiter, 1987]. The original justification for this approach was the that people often "jump to conclusions" in the context of deductively incomplete data.

Probabilists have noted some fundamental problems with these defeasible logics, which can lead them to jump to highly improbable conclusions. Most of them, for instance, are subject to some form of the lottery paradox. As a theory of inference, therefore, defeasible logics leave much to be desired.

Despite such problems, however, there are some domains where a defeasible logic may be an appropriate inference policy. Consider, for instance, a domain that satisfies the following criteria.

   *Intentionally Benign*. Inferential cues are intentionally designed to support correct inferences, particularly when negative consequences may result from false infernces.

   *Reliable Feedback*. If the agent acts in accordance with a false inference that may lead to a negative outcome, then the agent will receive feedback that the inference was false.

   *Opportunity to Backtrack*. The agent will have an opportunity to backtrack decisions prior to the occurrence of significant negative consequences.

In a consistently benign environment categorical inferences based on a defeasible logic seems an appropriate inference policy, even though the logic itself may be inappropriate as a theory of inference.

226

## 3.0 NONSTANDARD INFERENCE POLICIES

We define a nonstandard inference policy to be an inference procedure that does not correspond to any seriously considered theory of inference found in the literature. Obviously, the classification of an inference policy as nonstandard may change with the promotion of new theories. In this section, we examine some possible nonstandard policies.

### 3.1 Ratios of Possibilities

Logical probability theory not withstanding, perhaps one of the most maligned concepts in inference theory is the idea that one can calculate a reasonable belief value for a proposition by deducing the ratio of possible states in which the proposition is true. To give a typical counter example, if we accept the axiom a-->b, then Bel(b)=.67; since {a,b}, {~a,b}, and {~a,~b} are the three possible states. Suppose, however, that a is "Rover is a brown dog." and b is "Rover is a dog." In that case, the axiom a-->b certainly does not add any evidence that should impact ones degree of belief that "Rover is a brown dog." Yet according to the Possibility Ratio approach it has a major impact. Clearly, therefore, there is no *necessary* connection between a ratio of possible states and the perceived probability of a proposition. Consequently, it is hard to imagine how a theory of inference can be based solely on possible world ratios.

However there may be domains where the simplistic ratio approach is an appropriate inference policy. This is because the procedure for enumerating possible states is rarely arbitrary. To see how this works, consider Laplace's rule of induction. This rule states that in a series of observations of some event $a$ or $~a$, that after observing N occurences of $a$, and no instances of $~a$, then the inductive probability that $a$ will occur on the next trial is 1+N/2+N. This rule of induction is a special case of Carnap's c* function, which in turn is one instance of a family of coherent induction functions [Carnap, 1952]. Now consider a truth table containing the sixteen possible states for four propositions: $a$, $b$, $c$ and $d$. The proposition of interest is $a$. The other propositions are considered as candidates for a deterministic causal model for predicting $a$. Initially no causal connections are posited. Consequently, the possibility ratio (PR) of $a$ is PR($a$)=1/2. After one observation of $a$ we posit the causal rule $b$-->$a$. Now $a$ will be contained in exactly 8 of the 12 remaining possible states; so PR($a$)=2/3. After event $a$ occurs again, we add $c$-->$b$ - giving us PR($a$)=3/4. Finally, after $a$ occurs a fourth time, we add $d$-->$c$, giving us PR($a$)=4/5. Continuing this process we see the PR(a)=1+N/2+N. Our learning mechanism replicated this rule of induction. In general, *any* coherent rule of induction can be emulated with a causal learning model [Lehner, 1989].

We now turn this around. If a causal learning scheme responds to new instances by seeking deterministic rules for predicting that

227

instance, then one would expect a postive correlation between the relative frequency of an event, and the proportion of possible states containing that event. The more often X occurs, the greater the number of factors perceived as causally leading to X, resulting in a greater proportion of logically possible states containing X.

For some domains, therefore, ratios of possible states may provide a perfectly reasonable inference policy. Even though the causal learning mechanism may not explicitly take into account probabilistic considerations (e.g., as in most concept learning and explanation-based learning systems), there may be good reason to believe one can extract reasonable belief values from such systems.

3.2 Possibility and Probability

In section 1.0, we discussed an inference domain where reliable judgments were required. Here we expand a little on this idea. Consider the following problem. An inference system must be developed that must service the information requirements of multiple decision systems. Each decision system will query the inference module as needed regarding the status (truth value or degree of support) of certain propositions. The specific propositions queried will vary in each context.

Since the propositions to be queried cannot be predicted, it is decided that the inference system will maintain an up-to-date description of the current situation. That is, for some set of atomic propositions and their logically distinct combinations, the system should be able to report a belief value on request. Finally, it is considered important that the inference system be reliable. That is, for each set $S_X$ (all sentences believed to degree X), the expected proportion of truths in $S_X$ is X. The reason for this is simply that from one problem to the next, the elements of $S_X$ that are queried is unpredictable (more or less "random"). Consequently, if the system is reliable then the expected proportion of truths of propositions *reported* with degree of belief X is X.

What type of inference policy would guarantee satisfying these requirements? As it turns out [Lehner, 1989], *provable reliability* is achievable only if the system maintains (A) a set of possible states that contain the true state, (B) a set (possibly empty) of reliable probability statements that assigns point-values to a partition of the possible states, and (C) belief values are set equal to

$$c(q|r_1)p(r_1) + \ldots + c(q|r_n)p(r_n),$$

where r1...rn are sentences uniquely defining each partition, $p(r_1)$ is the probability of $r_1$, and $c(q|r_1)$ is the ratio of possible states in the $r_1$-partition that contain q. Furthermore, *precise reliability* (for each set $S_X$ *exactly* X proportion are true) can always be achieved by ignoring all probability information and only the possible states ratio.

This result has an interesting ramification. Reliability is always achievable, but only at the cost of ignoring some useful probability information. Reliability and accuracy tradeoff. Minimizing expected error requires conformance to the probability calculus, thereby giving up on reliability. On the other hand, reliability is only guaranteed if the inference system reports judgments that do not conform to the probability calculus. To illustrate, suppose an inference system knew $p(a)=.8$ and $p(b)=.6$, but had no information on $p(a\&b)$. As shown in Table 1, there are two sets of belief values that are provably reliable, and one that is precisely reliable.

TABLE 1
ILLUSTRATION OF THE TRADEOFF BETWEEN ACCURACY AND RELIABILITY
(expected error = $p(a)[1-b(a)]^2 + p(\sim a)[0-b(a)]^2 + p(b)[1-b(b)]^2 + p(\sim b)[0-b(b)]^2$ )

| Belief Values a | b | Expected Error (for a and b only) | Reliability (for all statements) |
|---|---|---|---|
| .8 | .6 | .4 | none guaranteed |
| .8 | .5 | .41 | provably reliable |
| .5 | .6 | .49 | provably reliable |
| .5 | .5 | .50 | precisely reliable |

From the perspective of inference policy, therefore, the appropriate degree of belief calculus depends on the relative importance of reliability vs. accuracy.

Note here how the characterization of an inference domain impacts the assessment of whether an inference policy is appropriate. The importance of provable reliability depends in part on the *inability* to anticipate which propositions will be queried. If we knew, for instance, that the inference domain was such that only propositions for which reliable probability information was available would be queried, then ignoring this probability infor-mation would make little sense.

### 3.3 Introspection and Probability

A concept endemic to nonmonotonic reasoning logics is the idea that negative introspection can provide evidential support for a hypothesis. For instance, in an autoepistemic logic, the sentence ~L~p-->p reads "If I cannot conclude ~p, p is true," or equivalently, "if p were false, I'd know it."

In everyday human affairs, this type of reasoning is quite common. It occurs whenever a person feels that he or she is knowledgeable on some topic ("My husband could not have been cheating on me," she said to the inspector, "for if he were, I would surely have known it.") It is also a characteristic of

229

most conversations, where by convention it is assumed that all relevant information is communicated [Reiter, 1987].

Probability Models of Negative Introspection

From a probability perspective, evidence-from-introspection provides some fascinating problems. In a probabilistic system, one could conceivable model categorical belief using an epsilon semantics [Pearl, 1988]. That is, if X is an agent's belief threshold, then the agent believes p (i.e., Lp) if $P(p|E)>X$, where E is the current evidence. An epsilon-semantics translation of $~L~p-->p$ might be $P(p|'P(p)>(1-X)')=X$.

If $X=.1$, then as long as the agent cannot deduce $~p$ with probability .9, that agent immediately concludes p with probability .9. This seems reasonable, if a little unusual. Suppose however that the agent decides to set a more conservative belief threshold, say $X=.99$. Now our agent concludes p with probability .99 whenever $~p$ cannot be deduced with .99 certainty. The more conservative the threshold, the less evidence needed for the agent to jump to a stronger conclusion. An epsilon semantics seems inappropriate here. Other self-referential approaches seem to have similar problems.

Given problems such as these a probabilist might be tempted to suggested that belief models should not have probability values conditional on self-referential probability statements, but should only be conditional on the original evidence items. A statement such as $~L~p-->p$ could simply be interpreted as $P(p|~E_1~...~~E_n) = High$, where the $E_i$ are relevant evidence items which did *not* occur. However, this approach fails to account for the fact that people *do* seem to use negative introspection as a source of evidence. Consequently, it cannot be used to encode human expert judgment. (As far as I know there is no reason to believe that negative introspection is inherently incoherent.) Also, the number of nonoccuring evidence items can be quite large, if not infinite -- making the development of such models infeasible in practice and perhaps impossible in theory.

Probability Analysis of Negative Introspection

Whether or not it is possible to develop probability model of negative introspection is unrelated to the issue of whether or not a probability *analysis* of negative introspection is useful. That is, a probabilistic analysis *of* an introspection-based inference policy may be quite informative.

To illustrate, consider the default rule $a:b|--b$, which states that if proposition a is believed and it is consistent to believe b, then infer b. The autoepistemic logic equivalent of this rule is $La \wedge ~L~b --> b$. Presumably, when a knowledge engineer adds a default rule like this to a knowledge base she believes that for the inference domain to which it will be applied the default rule will usually generate a valid conclusion. As a result whether or not an inference system implements a probability model, there is still a *probabilistic justification* for each default rule that is added to a knowledge base. Probabilistically, the standard justfication for a rule such as this is simply that $P(b|a)$=High, while an alternative justification, based on the communication convention interpretation, might be $P(b|La \char`\^ \char`\~ L \char`\~ b)$=High.

Consider the following case. An inference system contains the default rules $\{a:b|-\!-b, c:d|-\!-d\}$ and material implications $\{d-\!->\char`\~b, c-\!->a\}$. Upon learning c, two extensions result, one containing b and ~d, the second containing ~b and d. If the rules are interpreted in the standard way, then the first rule can be shown to be *provably irrelevant* since $P(b|c)=P(b|a \char`\^ c) \leq (1 - P(d|a \char`\^ c))=P(d|c)$, where by provable irrelevance I simply mean that enough evidence has been acquired to make the posterior assessment of the probability of b independent of the value of $P(b|a)$ in any fully specified probability model. If in fact the knowledge engineer had in mind the standard probability justifications for her default rules, then the default logic, by generating two extensions, is behaving in a manner inconsistent with the intentions of the knowledge engineer. Such a system does not reflect a satisfactory inference policy.

On the other hand, if it is assumed that default rules reflect communication conventions, then the alternative form for the probability justfications more closely reflects the knowledge engineers beliefs about the inference domain. In this case, however, $a:b|-\!-b$ cannot be shown to be provably irrelevant since $P(b|Lc \char`\^ La \char`\^ \char`\~ L \char`\~ b) = P(b|Lc \char`\^ \char`\~ L \char`\~ b)$, of which nothing can be derived. More generally, if negative introspection on categorical beliefs is viewed as a source of evidence for a default conclusion, then *no* extension can be anomalous in the sense that the probability justification for an applicable rule can never be shown to be provably irrelevant to a current problem. However, nonmonotonic logic theorists seem greatly concerned with the anomalous extension problem [Morris, 1988] suggesting therefore that nonmonotonic reasoning cannot be justified solely by the notion of communication conventions.

## 4.0 SUMMARY AND DISCUSSION

In this paper, an approach to inferencing under uncertainty was explored that calls for the specification of inference policies tailored to specific inference domains. Although the approach seems pluralistic, I claim no conflict with the Bayesian viewpoint that a rational/coherent system of belief values should conform to the probability calculus. As a good scientist, I find the objective of minimizing the error rate of my theories very compelling. Furthermore, my theories involve the development of algorithms that I hope will usually work. Consequently, I feel compelled to reason probabilistically about the relative frequency that applications of my theories will "work". However, in my (hopefully) cohe-rent reasoning *about* inference domains I can envision domains where global non-additive objectives (e.g., global reliability) are desirable. Consequently, I see no reason why coherent reasoning about an inference domain should necessarily lead to Bayesian inference policy as the preferred approach to inferencing *within* a domain.

## ACKNOWLEDGEMENTS

I would like to thank Dave Schum and two anonymous reviewers for their comments on an earlier draft of this paper.